\documentclass[10pt,twocolumn,letterpaper]{article}

\usepackage[review]{cvpr}      % To produce the REVIEW version
\definecolor{cvprblue}{rgb}{0.21,0.49,0.74}
\usepackage[pagebackref,breaklinks,colorlinks,allcolors=cvprblue]{hyperref}

\title{NeRF Inpainting with Geometric Diffusion Prior and Balanced Score Distillation}

\author{
  Menglin Zhang~~~~Xin Luo~~~~Yunwei Lan~~~~Chang Liu~~~~Rui Li~~~~Kaidong Zhang~~~~Ganlin Yang~~~~Dong Liu\\
  University of Science and Technology of China, Hefei, China\\
  {\tt\small \{zhangmenglin, xinluo, ywlan\}@mail.ustc.edu.cn, \tt\small dongeliu@ustc.edu.cn} \\
  \small\textbf{\url{https://github.com/Arcxml/GB-NeRF}}  % 替换为你的 GitHub 链接
}
\usepackage{hyperref}

\renewcommand{\sectionautorefname}{\S}

\begin{document}
% \maketitle

\twocolumn[{%
\renewcommand\twocolumn[1][]{#1}%

\maketitle
\begin{center}
    \centering
    \includegraphics[width=\linewidth]{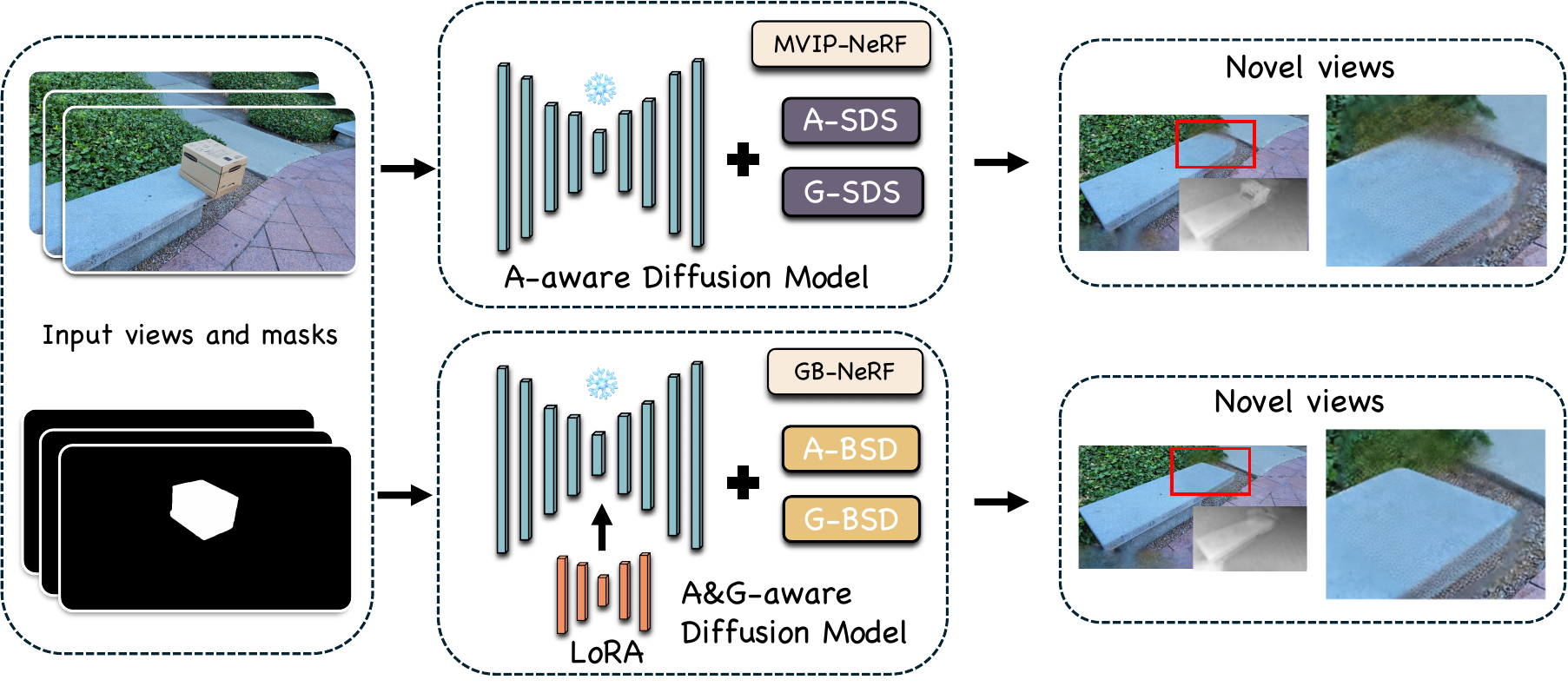}
    \captionof{figure}{Overview of our GB-NeRF framework compared to MVIP-NeRF. Both approaches leverage appearance (A) and geometric (G) priors from diffusion models through score distillation. To enhance geometric accuracy, we introduce two key innovations: (1) a specialized fine-tuning strategy using RGB-normal image pairs; (2) Balanced Score Distillation (BSD), which eliminates high-variability terms present in existing methods like SDS~\cite{poole2022dreamfusion} and CSD~\cite{yu2023text}, providing more stable supervision for occluded regions. Compared to MVIP-NeRF~\cite{MVIPNeRF}, our method achieves superior consistency and accuracy in inpainted regions.}\label{fig:teaser}
\end{center}%
}]

\begin{abstract}
% While effective NeRF inpainting methods have been developed utilizing explicit RGB and depth 2D inpainting supervision, their efficacy is fundamentally constrained by the limitations of the underlying 2D inpainting techniques. This is primarily reflected in two aspects: first, the pre-trained diffusion priors inadequately capture geometric features; second, existing methods related to Score Distillation Sampling (SDS) do not accurately capture the 3D distribution.

% To address these limitations, we propose a novel approach termed NN-NeRF, which modifies diffusion priors for NeRF inpainting, with the objective of enhancing both appearance and geometric fidelity. We have devised an effective fine-tuning method for diffusion that enables the original diffusion model to generate high-quality images while concurrently producing high-quality geometric priors. This allows us to obtain normal maps as a geometric representation and introduce a normal loss to incentivize accurate geometry inpainting and alignment with the visual appearance.Furthermore, we have identified potential shortcomings in Score Distillation Sampling (SDS) and its enhanced variant, Conditional Score Distillation (CSD). In response, we present our own method, Normalized Score Distillation (NSD). Our approach not only achieves superior quality but also enhances efficiency. Experimental results substantiate that our method significantly improves the recovery of both appearance and geometry compared to previous NeRF inpainting techniques.

Recent advances in NeRF inpainting have leveraged pretrained diffusion models to enhance performance. However, these methods often yield suboptimal results due to their ineffective utilization of 2D diffusion priors. The limitations manifest in two critical aspects: the inadequate capture of geometric information by pretrained diffusion models and the suboptimal guidance provided by existing Score Distillation Sampling (SDS) methods. To address these problems, we introduce GB-NeRF, a novel framework that enhances NeRF inpainting through improved utilization of 2D diffusion priors. Our approach incorporates two key innovations: a fine-tuning strategy that simultaneously learns appearance and geometric priors and a specialized normal distillation loss that integrates these geometric priors into NeRF inpainting. We propose a technique called Balanced Score Distillation (BSD) that surpasses existing methods such as Score Distillation (SDS) and the improved version, Conditional Score Distillation (CSD). BSD offers improved inpainting quality in appearance and geometric aspects. Extensive experiments show that our method provides superior appearance fidelity and geometric consistency compared to existing approaches.

\end{abstract}    
\section{Introduction}
\label{sec:intro}

As a pioneering work in neural rendering~\cite{henzler2020learning,niemeyer2020differentiable,MildenhallSTBRN20}, NeRF (Neural Radiance Field)~\cite{MildenhallSTBRN20} reconstructs complete 3D scenes from partial viewpoint observations, enabling exceptional 3D reconstruction and novel view synthesis. Practical applications often involve missing areas or unwanted objects that need to be removed. This creates a dual challenge: rendering unobserved viewpoints and filling in missing parts. NeRF inpainting addresses these challenges by reconstructing complete 3D scenes from masked images, ultimately resulting in a full NeRF model of the scene. This technology applies to various areas of 3D content creation, including object removal and scene completion. In this study, we will specifically focus on the task of removing objects.

Traditional NeRF inpainting methods generally follow a two-stage process: first, using 2D inpainting models to complete each view independently, and then employing these inpainted images to train the NeRF model~\cite{mirzaei2023spin}. However, these methods often yield inferior results due to the limited performance of 2D models and a lack of 3D consistency. The emergence of diffusion models~\cite{sohl2015deep,ho2020denoising,rombach2022high} has introduced promising solutions. DreamFusion~\cite{poole2022dreamfusion} pioneered Score Distillation Sampling (SDS) to leverage 2D diffusion image priors for 3D scene generation. Building on this, MVIP-NeRF employs complementary score distillation losses to enhance both appearance and geometry reconstruction. Despite these advancements, current methods struggle to generate high-quality NeRFs with accurate geometry due to the ineffective use of 2D diffusion priors, which arise from inadequate geometric information capture and suboptimal optimization guidance.

To address these challenges, we introduce GB-NeRF, a novel framework that leverages 2D diffusion priors for high-quality NeRF inpainting. We develop a fine-tuning strategy that trains the model to generate both RGB images and normal maps, utilizing a high-quality RGB-normal image dataset enhanced with captions generated by BLIP.~\cite{li2022blip}. We also incorporate LoRA~\cite{hu2022lora} into the U-Net~\cite{ronneberger2015u} and text encoder to learn appearance and geometric information. Additionally, we observe that existing score distillation methods, such as SDS~\cite{poole2022dreamfusion} and CSD~\cite{yu2023text}, suffer from unnecessary optimization variability due to random noise and unconditional noise prediction terms. This variability presents a challenge for NeRF inpainting tasks, as consistent supervisory signals are crucial in occluded regions. To mitigate this, we introduce Balanced Score Distillation (BSD), designed specifically for NeRF inpainting tasks, which reduces optimization uncertainty and improves quality in appearance and geometric aspects.

We perform extensive experiments on two representative datasets: \textit{LLFF}~\cite{mildenhall2019llff}, a conventional dataset, and \textit{SPIn-NeRF}~\cite{mirzaei2023spin}, a more challenging dataset. The results of our experiments demonstrate that our method achieves state-of-the-art performance in both quantitative metrics and visual quality by effectively utilizing diffusion priors. Our primary contributions can be summarized as follows:
\begin{enumerate}
    \item We propose a specialized diffusion model fine-tuning strategy that improves geometric understanding by learning to generate both RGB images and normal maps together, leading to more effective geometric priors for NeRF inpainting.
    \item We introduce Balanced Score Distillation (BSD), a novel optimization technique specifically designed for NeRF inpainting. By removing high-variability terms, BSD offers more stable and consistent supervision signals, enhancing optimization efficiency and improving reconstruction quality.
    \item We conduct comprehensive experiments using two representative datasets, demonstrating that our method achieves state-of-the-art performance in both visual quality and geometric accuracy for NeRF inpainting tasks.
\end{enumerate}

%-------------------------------------------------------------------------

\begin{figure*}[t]
  \centering
  \includegraphics[width=\linewidth]{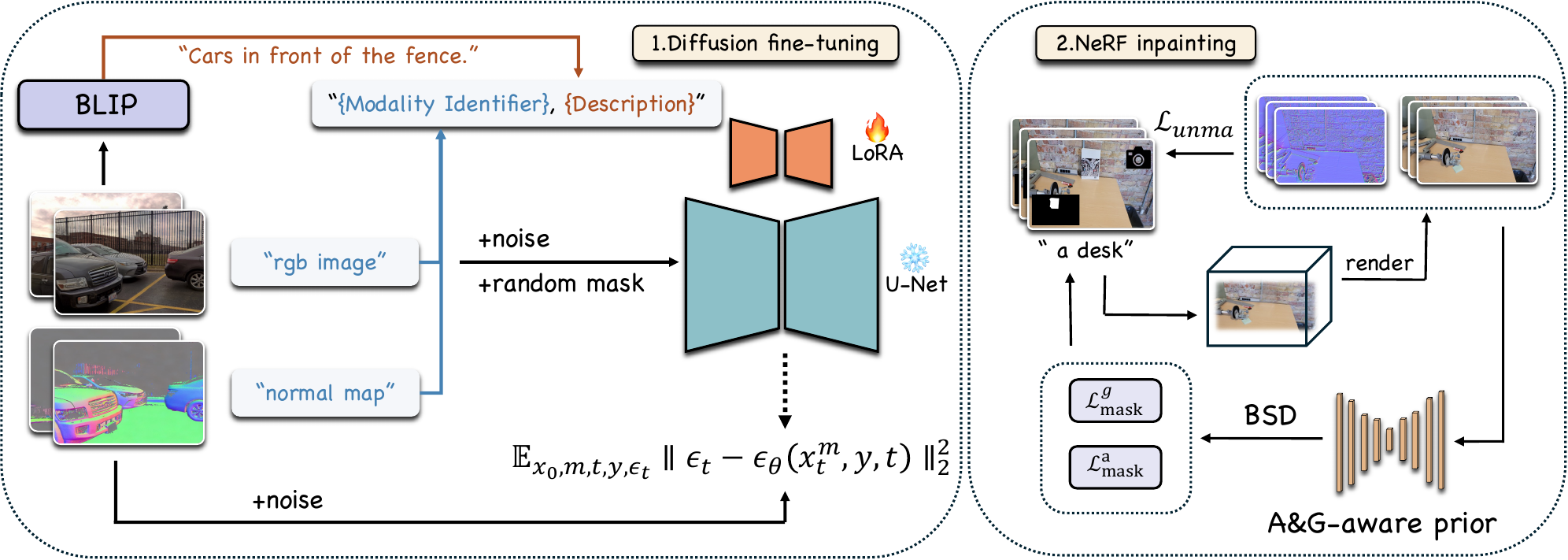}
\caption{Method overview. (Left) Diffusion fine-tuning: Our approach utilizes the DIODE dataset~\cite{diode_dataset}, which provides high-quality RGB images and corresponding normal maps. Captions are generated from RGB images using BLIP~\cite{li2022blip} and shared with their corresponding normal maps to leverage Stable Diffusion's text understanding capabilities. Modality identifiers (`normal map' or `RGB image') are prepended to these captions to distinguish between modalities. LoRA is integrated into both U-Net and text encoder to enhance the model's learning capacity further. (Right) NeRF inpainting: Given posed RGB images with corresponding masks and text descriptions, GB-NeRF reconstructs realistic textures and accurate geometry through dual supervision. In unmasked regions, direct pixel-wise RGB reconstruction loss ($L_{unma}$) provides supervision, while in masked areas, our BSD loss guides both RGB image and normal map generation using the fine-tuned diffusion model.}
  \label{fig:pipeline}
\end{figure*}
\renewcommand{\sectionautorefname}{\S}
\section{Related Work}
\subsection{NeRF Inpainting}
Recent research on NeRF inpainting has primarily focused on addressing 3D inconsistency using restored individual 2D images. Existing studies introduce methods to modify objects represented by NeRFs, such as EditNeRF~\cite{liu2021editing}, Clip-NeRF~\cite{wang2022clip}, and LaTeRF~\cite{mirzaei2022laterf}. A notable advancement is SPIn-NeRF~\cite{mirzaei2023spin}, which replaces traditional pixel loss with a more relaxed perceptual loss, enabling NeRF to reveal more high-frequency details while using depth predictions to supervise geometric structure. Another study utilizes LaMa for image inpainting, but its overall quality remains inferior to diffusion models. InpaintNeRF360~\cite{wang2023inpaintnerf360} employs a similar strategy, relying on perceptual and depth losses to enhance appearance and shape. However, we found that relying solely on perceptual loss does not fundamentally resolve the issues, often leading to subpar results. In contrast, MVIP-NeRF~\cite{MVIPNeRF} leverages diffusion models as priors to improve appearance and geometric fidelity. We discovered that pretrained diffusion models inadequately capture geometric information, prompting us to fine-tune the model to fully utilize this information.

\subsection{Learning 3D model via Diffusion Priors}
Recently, significant advancements in image generation have been driven by diffusion models~\cite{Diff15, DDPM, ScoreBased, DiffBeatGAN}. By training on large-scale text-image pairs, these models have achieved remarkable success in text-to-image generation~\cite{Schuhmann2022LAION5BAO}, with Stable Diffusion as a notable example. Consequently, many efforts have explored using diffusion models as priors for various image restoration tasks.
In addition to 2D applications, diffusion priors have gained attention in 3D generation. A pioneering work is Dreamfusion~\cite{poole2022dreamfusion}, which utilizes multi-view 2D diffusion priors for 3D generation through SDS loss. SDS has been widely adopted in subsequent works~\cite{lin2023magic3d,metzer2023latent,chen2023fantasia3d,wang2023prolificdreamer,huang2023dreamtime,zhu2023hifa,shi2023mvdream} aimed at enhancing DreamFusion. For example, Magic3D~\cite{lin2023magic3d} and Fantasia3D~\cite{chen2023fantasia3d} explore optimizing mesh topology for efficient high-resolution rendering. Several methods~\cite{haque2023instruct,shao2023control4d,zhuang2023dreameditor} have also applied SDS to inpainting undesired regions in NeRF scenes, including Nerfiller~\cite{weber2023nerfiller} and MVIP-NeRF~\cite{MVIPNeRF}. However, these SDS-based methods often struggle to produce high-quality objects. Approaches like CSD~\cite{yu2023text} and VSD~\cite{wang2023prolificdreamer}, which focus on distillation sampling, still face optimization variability due to random noise and unconditional noise prediction terms. This variability poses challenges for NeRF inpainting tasks, where consistent supervision signals in occluded regions are crucial. To address this, we enhanced the distillation method to provide stable supervision signals in occluded areas, improving generation quality.

%-----------------------------------------------------------------
\section{Preliminary}
\noindent\textbf{Neural Radiance Fields.} NeRFs~\cite{MildenhallSTBRN20} represent a 3D scene using a function $g$ that maps a 3D coordinate $\mathbf{p}$ and a viewing direction $\mathbf{d}$, to a color value $\mathbf{c}$ and a density $\sigma$. Specifically, the function $g$ is a neural network parameterized by $\theta$, where $g_{\theta}:(\gamma(\mathbf{p}),\gamma(\mathbf{d}))\mapsto(\mathbf{c}, \sigma)$, with $\gamma$ denoting the positional encoding. Each expected pixel color $\hat{C}(\mathbf{r})$ is rendered by casting a ray $\mathbf{r}$ with near and far bounds $t_n$ and $t_f$. The ray segment is typically divided into $N$ intervals ($t_1, t_2, ..., t_N$), and the pixel color is computed by $\hat{C}(\mathbf{r})=\sum_{i=1}^{N}w_i\mathbf{c}_i$, where $w_i=T_i(1-\mathrm{exp}(-\sigma_i\delta_i))$, $T_i=\mathrm{exp}(-\sum_{j=1}^{i-1}\sigma_j\delta_j)$, and $\delta_i=t_{i}-t_{i-1}$. Mathematically, the NeRF reconstruction loss is formulated as:

\begin{equation} \label{eq:reco_a} \mathcal{L}^{a}=\sum_{\mathbf{r}\in R}||\hat{C}(\mathbf{r})-C(\mathbf{r})||^{2}, \end{equation}
\noindent where $\hat{C}(\mathbf{r})$ is the rendered pixel color from the $N$ samples, $R$ is the batch of rays sampled from the training views, and $C(\mathbf{r})$ is the ground truth color for the pixel. When depth information is available, an additional reconstruction loss can be added to optimize the geometry of NeRF scenes~\cite{deng2022depth}:

\begin{equation} \label{eq:reco_g} \mathcal{L}^{g} = \sum_{\mathbf{r}\in R}||\hat{D}(\mathbf{r})-D(\mathbf{r})||^{2}, \end{equation}

\noindent where $\hat{D}(\mathbf{r})$ is the rendered depth, and $D(\mathbf{r})$ is the ground-truth depth for the pixel.

\noindent\textbf{Score Distillation Sampling.}
SDS~\cite{poole2022dreamfusion} allows the optimization of any differentiable image generator, such as NeRFs or images. Formally, let $\mathbf{x}=g(\theta)$ represent an image rendered by a differentiable generator $g$ with parameter $\theta$. SDS minimizes the density distillation loss~\cite{oord2018parallel}, which is essential to the KL divergence between the posterior of $\mathbf{x} = g(\theta)$ and the text-conditional density $p_\phi^{\omega}$:

\begin{equation} 
\begin{split}
\mathcal{L}_{\tt{Dist}}(\theta) =\mathbb{E}_{t,\boldsymbol{\epsilon}}\big[w(t)\,D_{\tt{KL}}\big( q\big(\mathbf{x}_t|\mathbf{x}\big) \,\|\, p_\phi^{\omega}(\mathbf{x}_t; y,t)\big) \big],
    \label{eq:KL}
\end{split}
\end{equation}

\noindent where $w(t)$ is a weighting function, $y$ is the text embedding, and $t$ is the noise level. For efficient computation, SDS updates the parameter $\theta$ by selecting random timesteps $t\sim \mathcal{U}(t_{\tt{min}}, t_{\tt{max}})$, forwarding $\mathbf{x}=g(\theta)$ with noise $\boldsymbol\epsilon\sim \mathcal{N}(\mathbf{0},\mathbf{I})$, and calculating the gradient as:

\begin{align}\label{eq:sds}
    \nabla_\theta \mathcal{L}_{\tt{SDS}}(\theta) = \mathbb{E}_{t,\boldsymbol{\epsilon}}\left[ w(t) \big(\boldsymbol{\epsilon}_\phi^{\omega}(\mathbf{x}_t; y,t) - \boldsymbol{\epsilon}\big)\frac{\partial \mathbf{x}}{\partial \theta} \right].
\end{align}

\section{Method}

Our task involves a set of RGB images, denoted as \( \mathcal{I} = \{ I_i \}_{i= 1}^{n} \), along with their corresponding camera poses \( \mathcal{G} = \{ G_i \}_{i= 1}^{n} \) and object masks \( \mathcal{M} = \{ m_i \}_{i= 1}^{n} \). The objective is to train a NeRF (Neural Radiance Fields) model to render inpainted scene content from any novel viewpoint. Our approach consists of two main components. First, we introduce our fine-tuning strategy, which allows diffusion models to learn both appearance and geometric priors (see Section \autoref{sec:geo.pri.enh}). Next, we enhance Classifier Score Distillation (CSD) (refer to Section \autoref{sec:csd.ref}) and propose Balanced Score Distillation (BSD), an improved alternative to traditional Score Distillation Sampling (SDS) methods (see Section \autoref{sec:bsd}). Figure~\ref{fig:pipeline} illustrates our entire approach.

\subsection{Geometric Prior Enhancement}\label{sec:geo.pri.enh}
Building on the insights from MVIP-NeRF~\cite{MVIPNeRF}, we understand the importance of accurate geometric reconstruction in NeRF inpainting. Although MVIP-NeRF has shown that pretrained Stable Diffusion has inherent geometric priors and can handle normal maps, our experiments indicate that it struggles to effectively complete these normal maps. This challenge arises mainly from diffusion models' tendency to produce overly smooth results, which can compromise the intricate geometric details crucial for accurate 3D reconstruction. To overcome this limitation, we fine-tune the diffusion model to improve its ability to generate structurally accurate normal maps while still being effective in generating RGB images.

% \subsubsection{Dataset Preparation}\label{sec:data.prep}
% Our proposed approach commences with a detailed training image processing methodology description. Initially, we construct the training dataset by selecting images from the DIODE dataset~\cite{diode_dataset} along with their corresponding normal maps. To enhance the learning capability of the diffusion model, we first employ BLIP~\cite{li2022blip} to generate specific captions for each RGB image. Since the diffusion model functions as a text-to-image framework, incorporating these captions is expected to significantly enhance its comprehension of textual information, thereby facilitating the generation of higher quality and more contextually relevant images. Since the normal maps are directly associated with the RGB images, they share identical captions. Subsequently, we process the captions to distinguish between normal maps and RGB images by adding prefixes such 

The fine-tuning process incorporates Low-Rank Adaptation (LoRA)~\cite{hu2022lora} into both the U-Net and text encoder, preserving Stable Diffusion's powerful image priors and text understanding capabilities. Our training data comes from the DIODE dataset~\cite{diode_dataset}, which provides RGB images and their corresponding normal maps. Besides, we use BLIP~\cite{li2022blip} to generate captions for RGB images, which are then shared with their corresponding normal maps. Each caption starts with a modality identifier—either ``normal map" or ``RGB image"—to distinguish between them, which is referred to as caption \(y\). The training process follows the methodology outlined in~\cite{zhuang2023task}, utilizing a self-supervised inpainting loss for both RGB images and normal maps. In this approach, random masks are sampled and combined for each training image.

\subsection{Balanced Score Distillation (BSD)}
Neural Radiance Fields (NeRFs) have significantly advanced the representation of 3D scenes, allowing for view synthesis from various angles. However, NeRFs face challenges in reconstructing masked or occluded regions within multi-view images. To address this issue, Score Distillation Sampling (SDS) and its improved version, Classifier Score Distillation (CSD), utilize pretrained diffusion models to enhance the optimization of NeRFs. This approach has captured our interest. 

However, we have observed that CSD struggles to achieve satisfactory results in NeRF inpainting tasks. We hypothesize that effective NeRF inpainting requires more stable supervision signals. To test this hypothesis, we introduce an additional hyperparameter to control the unconditional noise prediction term in CSD's formulation. Our analysis of the relationship between unconditional noise prediction and inpainting quality informs the development of an improved optimization strategy.

\subsubsection{Classifier Score Distillation (CSD) Refinement}\label{sec:csd.ref}

% For clarity, we focus on the diffusion-related terms in the optimization process. In SDS, the final gradient to update $\theta$ is expressed as:

% \begin{equation}
% \nabla_\theta \mathcal{L}_{\text{SDS}} = \mathbb{E}_{t, \epsilon, \mathbf{c}} \left[ w(t) (\epsilon_{\phi}(\mathbf{x}_t ;y,t) - \epsilon) 
% \frac{\partial \mathbf{x}}{\partial \theta} \right].
% \label{eq:sds}
% \end{equation}

% % the basic optimization gradient is given by 
% % \begin{equation}
% % \boldsymbol{\epsilon}_\phi^{\omega}(\mathbf{x}_t; y,t) - \boldsymbol{\epsilon}
% % \end{equation}
% A crucial aspect of score distillation involves computing the gradient to be applied to the rendered image $\mathbf{x}$ during optimization, which we denote as $\delta_x(\mathbf{x}_t ;y,t) := \epsilon_{\phi}(\mathbf{x}_t ;y,t) - \epsilon$. This formulation encourages the rendered images to concentrate in high-density areas that are conditioned on the text prompt \(y\). In practice, however, classifier-free guidance (CFG) is utilized in diffusion models with a large guidance weight $\omega$ (e.g., $\omega=100$ in DreamFusion \cite{poole2022dreamfusion}) to achieve high-quality results, causing the final gradient applied to the rendered image to deviate from \ref{eq:sds}. 

To better understand the limitations of existing methods, we examine the core optimization mechanism in score distillation. Specifically, the gradient applied to the rendered image $\mathbf{x}$ during optimization plays a crucial role. In SDS, this gradient is denoted as $\delta_x(\mathbf{x}_t ;y,t) := \epsilon_{\phi}(\mathbf{x}_t ;y,t) - \epsilon$. This formulation encourages the rendered images to concentrate in high-density areas that are conditioned on the text prompt \(y\). In practice, however, classifier-free guidance (CFG) is utilized in diffusion models with a large guidance weight $\omega$ (e.g., $\omega=100$ in DreamFusion \cite{poole2022dreamfusion}) to achieve high-quality results, causing the final gradient applied to the rendered image to deviate from \ref{eq:sds}.
Specifically, with CFG, $\delta_x$ is expressed as:

% \vspace{-1em}
\begin{equation}
\begin{aligned}
\delta_x(\mathbf{x}_t ;y,t) & = \underbrace{\left[\epsilon_{\phi}(\mathbf{x}_t;y,t) - \epsilon\right]}_{\delta_{x}^{\text{gen}}}\\ & +\omega\cdot\underbrace{\left[\epsilon_{\phi}(\mathbf{x}_t ;y,t)-\epsilon_{\phi}(\mathbf{x}_t;t)\right]}_{\delta_{x}^{\text{cls}}}. 
\label{eq:decompose}
\end{aligned}
\end{equation}
% \vspace{-1em}

% To ensure successful optimization, SDS employs classifier-free guidance with large guidance scales, transforming the actual optimization direction as follows: 
% \begin{equation}
% \begin{split}
% \ \left[\epsilon_\phi\left(\mathbf{x}_t ; y, t\right)-\epsilon\right]+\omega \cdot\left[\epsilon_\phi\left(\mathbf{x}_t ; y, t\right)-\epsilon_\phi\left(\mathbf{x}_t ; t\right)\right]
% \end{split}
% \end{equation}
Previous research~\cite{yu2023text} has demonstrated that the term $\delta_{x}^{\text{cls}}$ plays a crucial role in determining the optimization direction. Based on this observation, CSD simplifies the process by concentrating exclusively on this dominant term. By also incorporating negative prompts, CSD utilizes the following gradient:

\begin{equation}
\begin{aligned}
\delta_x^\text{cls} & = \omega_1 \cdot \epsilon_\phi\left(\mathbf{x}_t ; y, t\right) + \left(\omega_2 - \omega_1\right) \cdot \epsilon_\phi\left(\mathbf{x}_t ; t\right) \\ &- \omega_2 \cdot \epsilon_\phi\left(\mathbf{x}_t ; y_{\text{neg}}, t\right),
\end{aligned}
% \end{split}
\end{equation}
\noindent where the unconditional noise prediction term $\epsilon_\phi\left(\mathbf{x}_t; t\right)$ in the equation is determined by two parameters simultaneously. To enhance its flexibility, we modify the formula as follows:
\begin{equation}
\begin{split}
\delta_x^\text{cls} & = \,\omega_1 \cdot \epsilon_\phi\left(\mathbf{x}_t ; y, t\right) \\ &  +\omega_3 \cdot \epsilon_\phi\left(\mathbf{x}_t ; t\right)-\omega_2 \cdot \epsilon_\phi\left(\mathbf{x}_t ; y_{\text {neg }}, t\right),
\end{split}
\end{equation}

\noindent which provides greater flexibility in controlling each term's contribution to overall optimization.

\begin{figure}[t]
    \centering
    \includegraphics[width=\linewidth]{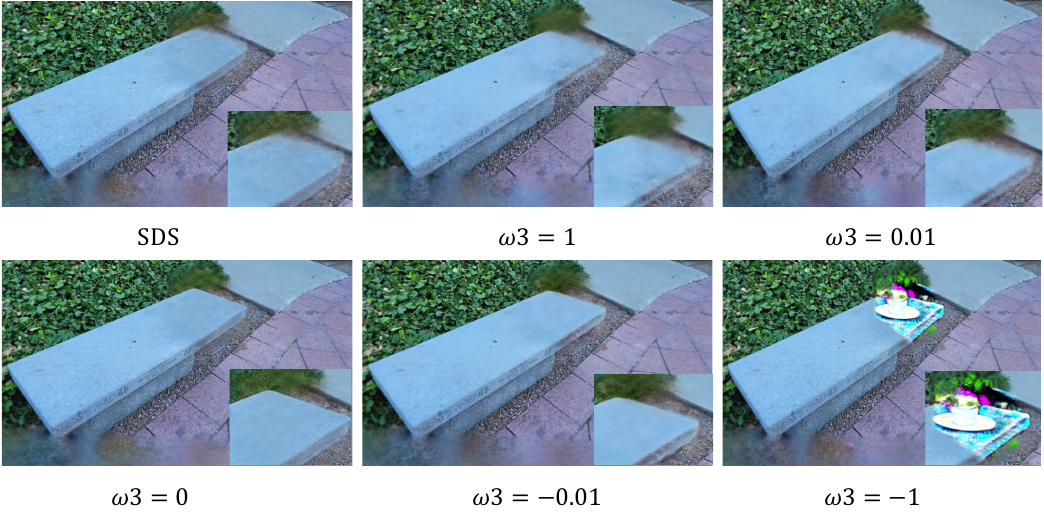}
    \caption{Impact of tuning coefficient $\omega_3$ on NeRF inpainting. The incorporation of unconditional noise prediction introduces excessive randomness, resulting in degraded inpainting quality.}
    \label{fig:w_comp}
\end{figure}

To investigate the impact of unconditional noise prediction on NeRF inpainting, we conduct experiments with varying values of $\omega_3$, as shown in Figure \ref{fig:w_comp}. Our results reveal that positive weights lead to blurry reconstructions, while negative weights introduce undesirable artifacts. Notably, the reconstruction quality significantly improves as $\omega_3$ approaches zero. This observation aligns with theoretical expectations, as unconditional predictions inherently exhibit higher diversity compared to conditional ones. These findings suggest that NeRF inpainting benefits from more deterministic score estimation, which motivates our proposed improvement detailed below.

\subsubsection{Balanced Score Distillation} \label{sec:bsd}
% To investigate the impact of unconditional noise prediction on NeRF inpainting, we conduct experiments with varying values of $\omega_3$, as shown in Figure \ref{fig:w_comp}. Our results reveal that positive weights lead to blurry reconstructions, while negative weights introduce undesirable artifacts. Notably, the reconstruction quality significantly improves as $\omega_3$ approaches zero. This observation aligns with theoretical expectations, as unconditional predictions inherently exhibit higher diversity compared to conditional ones. These findings suggest that NeRF inpainting benefits from more deterministic score estimation, which motivates our proposed improvement detailed below.

Our analysis confirms consistent and stable supervision signals in occluded regions are crucial for high-quality NeRF inpainting. Based on this insight, we propose to simply eliminate the unconditional noise prediction term, resulting in the following formulation:
\begin{equation}
\begin{split}
\delta_x^\text{BSD} = & \, \omega_1 \cdot \epsilon_\phi\left(\mathbf{x}_t ; y, t\right) - \omega_2 \cdot \epsilon\phi\left(\mathbf{x}_t ; y_{\text{neg}}, t\right).
\end{split}
\end{equation}

We refer to this approach as Balanced Score Distillation (BSD) because it ensures a balance between positive and negative prompts. BSD simplifies CSD while achieving superior performance, requiring only two network evaluations instead of three, which reduces computational overhead and simplifies parameter tuning. Compared to SDS, our approach eliminates the random noise term and introduces conditional guidance with negative prompts. This design offers enhanced performance while maintaining algorithmic simplicity: the positive prompt term guides optimization toward desired outcomes, while the negative prompt term steers it away from undesirable results, making BSD particularly effective for NeRF inpainting tasks.

\subsubsection{Overall Loss}

Specifically, we utilize pixel-wise color (Eq.~\ref{eq:reco_a}) and depth (Eq.~\ref{eq:reco_g}) reconstruction loss for unmasked regions, which we denote as $\mathcal{L}_{\mathrm{unma}}^{a}$ and $\mathcal{L}_{\mathrm{unma}}^{g}$. For masked regions, we first render an RGB image and a normal map from the NeRF scene. We employ our BSD losses to compute a gradient direction iteratively for detailed and high-quality appearance and geometry completion within the mask. The loss in the mask of appearance can be written as:
\begin{equation}\label{eq:appearance_sds}
\begin{split}
    & \nabla_\theta \mathcal{L}_{\mathrm{masked}}^{a} = \\
    & \left[ \, \omega_1 \cdot \epsilon_\phi\left(\mathbf{x}_t ; y, t\right)  - \omega_2 \cdot \epsilon_\phi\left(\mathbf{x}_t ; y_{\text{neg}}, t\right) \right]\frac{
    \partial \mathbf{x}_t}{\partial \mathbf{x}}\frac{
    \partial \mathbf{x}}{\partial \theta}.
\end{split}
\end{equation}  

Similarly, the loss in the mask of geometry can be written as:
\begin{figure*}[t]
  \centering
  \includegraphics[width=\linewidth]{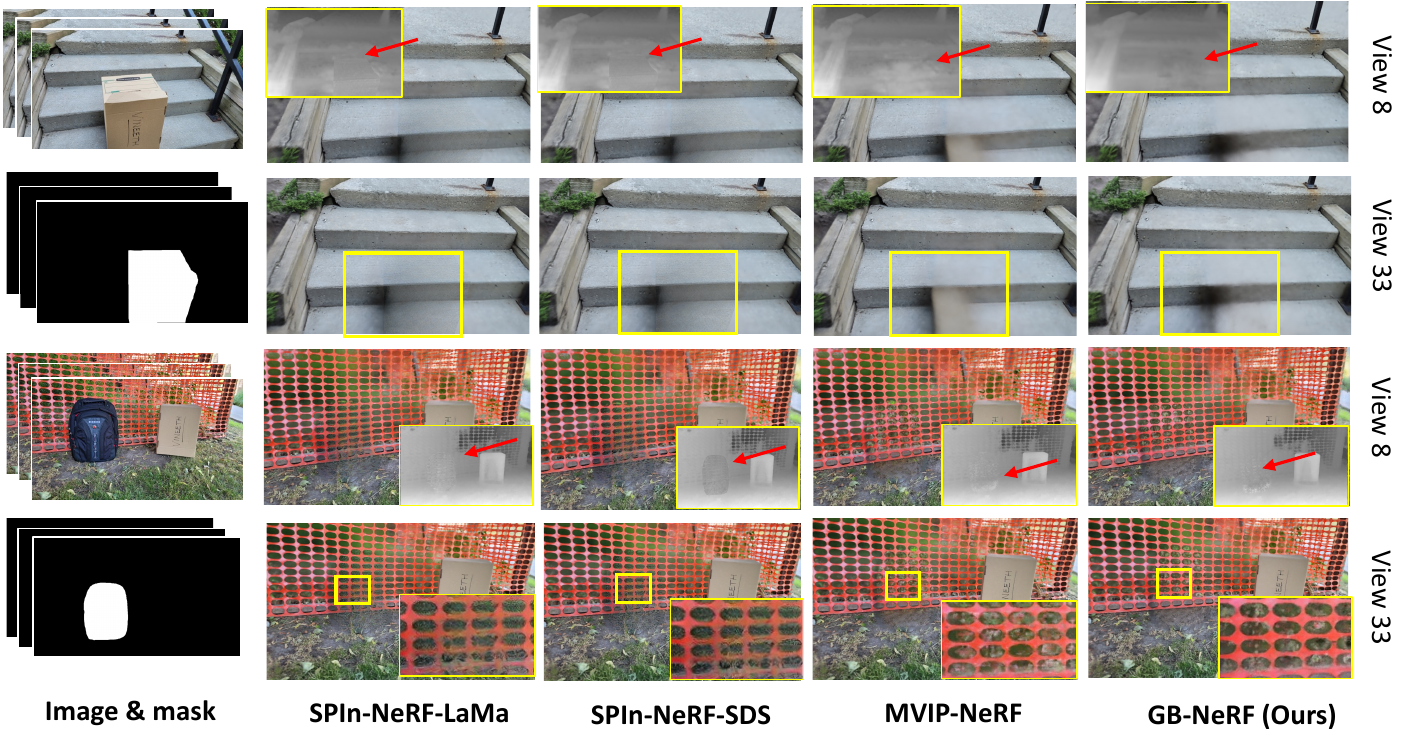}
\caption{Visual comparison with three representative approaches on two scenes. The first scene uses the prompt `A stair' while the second uses `A fence'. Our method effectively handles both scenarios, producing view-consistent results with superior geometric accuracy (note the well-preserved stair structure and depth, while (a) and (b) preserve residual geometry from original objects, and (c) generates artifacts) and realistic textures (observe the cleaner and more structured fence pattern in our results).}
  \label{fig:results_example1}
\end{figure*}

\begin{figure}[t]
    \centering
    
    \includegraphics[width=\linewidth]{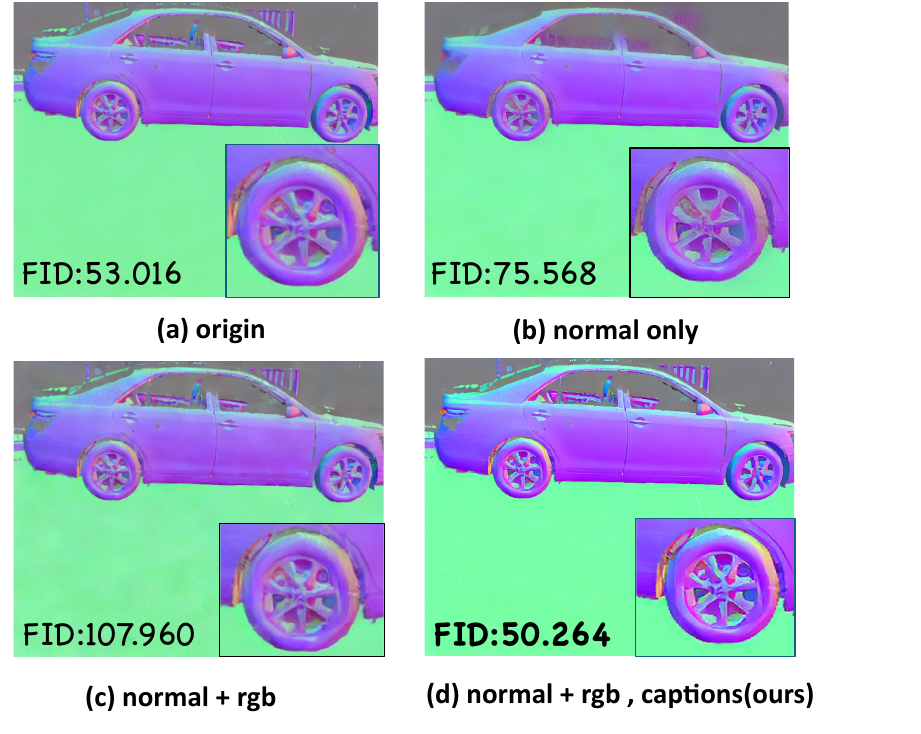}
    \caption{Comparison of different fine-tuning strategies for the diffusion model. (a) Original diffusion model without fine-tuning; (b) Fine-tuning with normal maps only, using modality identifier `normal map' as prompt; (c) Fine-tuning with both RGB images and normal maps, using modality identifiers `RGB image' and `normal map' as prompts; (d) Our approach: fine-tuning with RGB-normal image pairs using BLIP-generated captions prepended with modality identifiers as prompts. Results show that while strategies (b) and (c) underperform compared to the original model, our method significantly enhances the model's capability in normal map reconstruction.}
    \label{fig:normal_comp}
\end{figure}

\begin{equation}\label{eq:geometry_sds}
\begin{split}
    & \nabla_\theta \mathcal{L}_{\mathrm{masked}}^{g} =  \\
    & \left[ \, \omega_1 \cdot \epsilon_\phi\left(\mathbf{\mathbf{n}}_t ; y, t\right)  - \omega_2 \cdot \epsilon_\phi\left(\mathbf{n}_t ; y_{\text{neg}}, t\right) \right]\frac{
    \partial \mathbf{n}_t}{\partial \mathbf{n}}\frac{
    \partial \mathbf{n}}{\partial \theta}.
\end{split}
\end{equation}
% \begin{equation}\label{eq:appearance_sds}
% \begin{split}
%     \nabla_\theta \mathcal{L}_{\mathrm{masked}}^{a} = \delta_{x}^{\text{BSD}}\frac{
%     \partial \mathbf{x}_t}{\partial \mathbf{x}}\frac{
%     \partial \mathbf{x}}{\partial \theta},
% \end{split}
% \end{equation}  

\begin{figure*}[t]
  \centering
  \includegraphics[width=\linewidth]{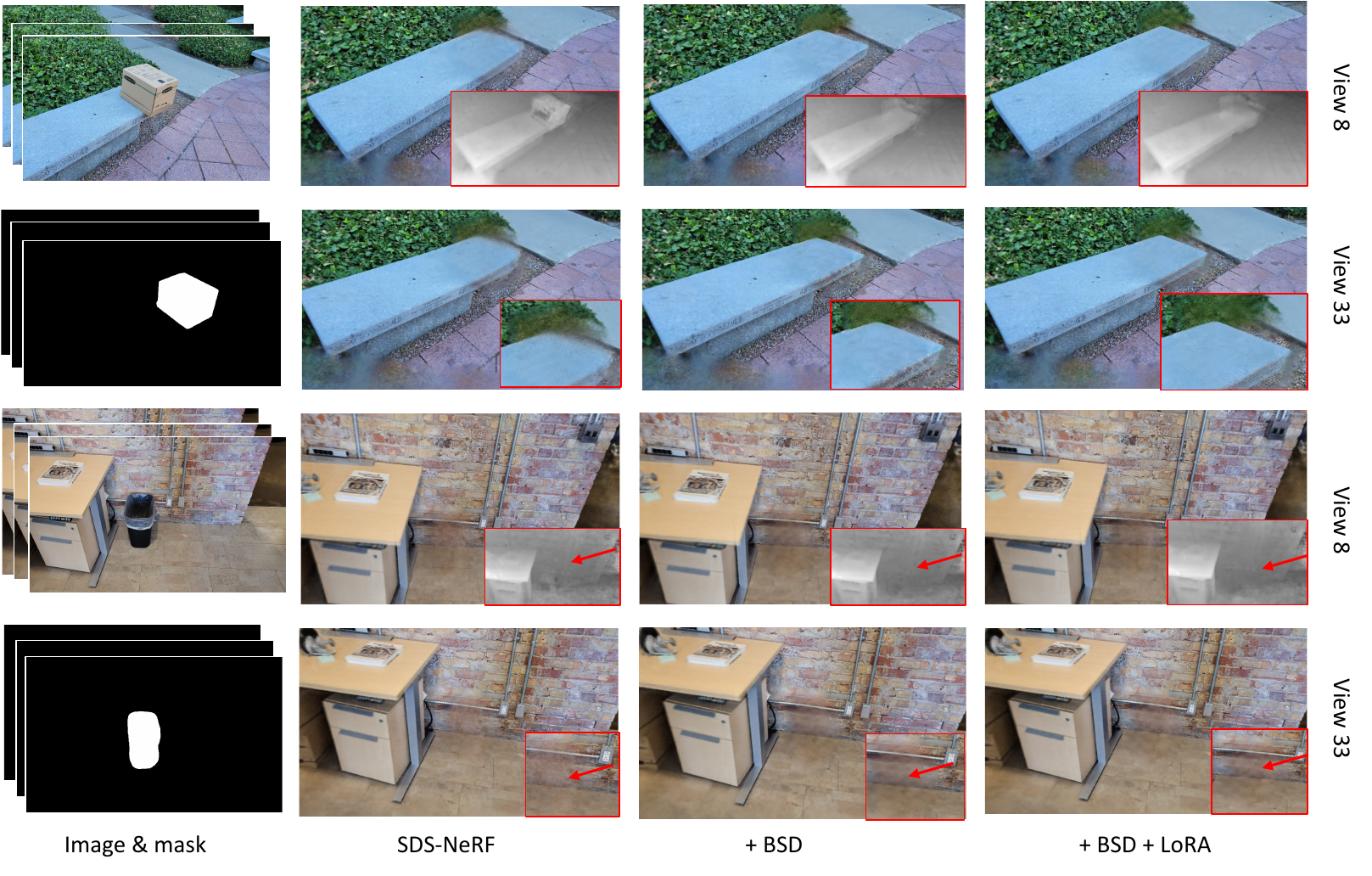}
\caption{Visual comparison of ablation study about each component. After adding BSD distillation, the blurriness around the edges is reduced (see the bench and the floor gaps), and further incorporating LoRA better reconstructs the geometry of the bench and the wall (see the depth map of the bench and floor).}
  \label{fig:ablation}
\end{figure*}

% \begin{equation}\label{eq:geometry_sds}
% \begin{split}
%     \nabla_\theta \mathcal{L}_{\mathrm{masked}}^{g} =  \delta_{\mathbf{n}}^{\text{BSD}}\frac{
%     \partial \mathbf{x}_t}{\partial \mathbf{n}}\frac{
%     \partial \mathbf{n}}{\partial \theta}.
% \end{split}
% \end{equation}
In general, we address unmasked and masked regions separately, following this general formulation:
% \begin{align}
% \label{eq:final_loss}
% \mathcal{L} &= \mathcal{L}_{\mathrm{unmasked}}^{a}  + \lambda_1\mathcal{L}_{\mathrm{unmasked}}^{g} \\
% &\quad + \lambda_2\mathcal{L}_{\mathrm{masked}}^{a} + \lambda_3\mathcal{L}_{\mathrm{masked}}^{g}.
% \end{align}

\begin{align}
\label{eq:final_loss}
\mathcal{L} &= \mathcal{L}_{\mathrm{unma}}^{a}  + \lambda_1\mathcal{L}_{\mathrm{unma}}^{g}  + \lambda_2\mathcal{L}_{\mathrm{mask}}^{a} + \lambda_3\mathcal{L}_{\mathrm{mask}}^{g}.
\end{align}

%\lambda_1\mathcal{L}_{\mathrm{unmasked}}^{g}

%-----------------------------------------------------------------

\begin{table*}[htbp]
\centering
\caption{\textbf{Quantitative comparison on \textit{SPIN-NeRF} and \textit{LLFF} datasets.} Our method demonstrates superior performance across most metrics compared to existing NeRF inpainting approaches.}
\resizebox{\textwidth}{!}{
\begin{tabular}{lcccccccccc}
    \toprule
    ~ & \multicolumn{8}{c}{\textit{SPIN-NeRF}} & \multicolumn{1}{c}{\textit{LLFF}} \\
     \cmidrule(lr){2-8} 
     \cmidrule(lr){9-11} 
    %\midrule
    ~ & PSNR$\uparrow$  & SSIM$\uparrow$ & FID$\downarrow$ & NIMA$\uparrow$ & BRISQUE$\downarrow$ & D-FID$\downarrow$ &  D-PSNR $\uparrow$ & FID$\downarrow$  & NIMA$\uparrow$ & BRISQUE$\downarrow$ \\
    \midrule 
    SPIn-NeRF + LaMa 
    & 19.651 & 0.4197 & 79.424 & 4.081 & 22.433 & 186.967 & 13.827 & 283.348 & 4.788 & 12.325 \\
    SPIn-NeRF + SDS 
    & 19.655 & 0.4199 & 79.196 & 4.101 & 22.420 & 191.496 & 13.834 & 286.742 & 4.729 & 12.345 \\
    MVIP-NeRF  
    & \textbf{19.813}  & 0.4208 & 72.616 & 4.455 & 23.562 & 172.127 &  13.914 & 285.873 & 4.862 &  11.901 \\
    GB-NeRF (Ours)
    & 19.489 & \textbf{0.4266} & \textbf{67.587} & \textbf{4.550} & \textbf{17.496} & \textbf{150.473} & \textbf{14.157} & \textbf{270.923}  & \textbf{4.915} & \textbf{11.711}\\
    \bottomrule
\end{tabular}}
\label{tab:baseline}
\end{table*}

\begin{table}[t]
    \centering
    \caption{\textbf{Ablation analysis.} Quantitative comparison of different model components. Enhanced geometric priors significantly improve geometric-related metrics (D-FID), while Balanced Score Distillation (BSD) enhances overall visual quality by reducing optimization uncertainty. The best and second-best results are highlighted in \textbf{bold} and \underline{underlined}, respectively.}
    \footnotesize
    \resizebox{\columnwidth}{!}{
        \begin{tabular}{l cccccc}
            \toprule
                      
            ~ & PSNR$\uparrow$  & SSIM$\uparrow$ & FID $\downarrow$  & NIMA $\uparrow$ & BRISQUE$\downarrow$ & D-FID $\downarrow$ 
             
            \\
            \midrule 
 origin 	& \textbf{19.814} & \underline{0.421} &72.616 & 4.455 & 23.562 & 172.127 
	\\
 +LoRA  & 19.543 & 0.402 & 70.137 & 4.432 & \textbf{17.029} & \textbf{147.330} 
	\\
 +BSD   & \underline{19.813} & 0.419 & \textbf{64.310} & \underline{4.550} & 23.002 &	161.416	
	\\
 +LoRA+BSD & 19.489 & \textbf{0.426}  & \underline{67.587} & \textbf{4.551} & \underline{17.496}  &	\underline{150.473}
	\\

            \bottomrule
        \end{tabular}}
    \label{tab:ablation}
\end{table}

\section{Experiments and Discussions }
\noindent \textbf{Implementation Details.} We implement our NeRF inpainting model based on SPIn-NeRF~\cite{mirzaei2023spin} and train it on a single NVIDIA A100 GPU for $10{,}000$ iterations using the Adam optimizer with a learning rate of $10^{-4}$.
For the diffusion prior, we set the size of all latent inputs to $256 \times 256$ and configure the timestep range with $t_{\texttt{min}} = 0.02$ and $t_{\texttt{max}} = 0.98$. For the loss of masked region, we set $\omega_1 = 7.5$ and $\omega_2 = 6.5$ for appearance BSD $\mathcal{L}_{\text{masked}}^{a}$, while using $\omega_1 = 1.5$ and $\omega_2 = 0.5$ for geometric BSD $\mathcal{L}_{\text{masked}}^{g}$. For the balance weights in the final loss function (Eq.~\ref{eq:final_loss}), we empirically set $\lambda_1 = 0.1$ and $\lambda_2 = \lambda_3 = 0.0001$. For diffusion fine-tuning, we set the learning rate to $10^{-4}$ and use LoRA of rank $32$.

\noindent \textbf{Datasets.} We evaluate our method on two real-world datasets: \textit{SPIn-NeRF} and \textit{LLFF}. Below, we provide detailed descriptions of these datasets.

\noindent \textit{SPIn-NeRF}~\cite{mirzaei2023spin} serves as an object removal benchmark consisting of $10$ scenes. For each scene, the dataset provides $60$ training views captured with an object intended for removal, accompanied by corresponding inpainting masks indicating the object's location. For evaluation purposes, we select $8$ challenging scenes from this dataset, where each scene includes $40$ testing views in which the target object has been physically removed during capture.

\noindent \textit{LLFF}~\cite{mildenhall2019llff} comprises multiple real-world scenes with varying numbers of images (ranging from $20$ to $45$). For our experiments, we utilize a four-scene subset that has been annotated with 3D grounded object removal masks. Since this dataset does not provide a separate test set, we follow previous methods and use the training views for evaluation purposes.
Following established practices in prior work~\cite{MVIPNeRF}, we standardize all images by resizing them to have a long-edge size of $1008$.

\noindent \textbf{Metrics.} To comprehensively evaluate our inpainting results, we use several complementary metrics. For direct comparison with ground-truth scenes, we utilize traditional full-reference metrics PSNR and SSIM, which measure pixel-level and structural similarities, respectively. To assess the perceptual quality of generated images, we employ no-reference metrics NIMA and BRISQUE, which evaluate image aesthetics and quality without requiring ground truth. Additionally, we use FID to measure the distribution distance between inpainting results and ground-truth scenes. For evaluating geometric accuracy, we employ D-PSNR to compare the reconstructed disparity maps with the ground-truth disparity maps. Additionally, we report D-FID to measure geometric consistency by calculating FID on the disparity maps.

\subsection{Results}
\noindent \textbf{Baselines.} Recent methods based on NeRF for inpainting have shown better performance than traditional image and video inpainting techniques. Therefore, we focus our comparison on state-of-the-art NeRF inpainting methods. Specifically, we compare against SPIn-NeRF~\cite{mirzaei2023spin}, SPIn-NeRF-SDS, and MVIP-NeRF~\cite{MVIPNeRF}. SPIn-NeRF applies the 2D inpainting model LaMa~\cite{suvorov2022resolution} to independently inpaint each image in the dataset before training NeRF with these inpainted images. SPIn-NeRF-SDS enhances the NeRF optimization process by applying additional supervision in masked regions using SDS. MVIP-NeRF improves geometric reconstruction by utilizing both an appearance SDS and a geometric SDS, which take RGB images and normal maps as input, respectively. To ensure a fair comparison, we use the official implementations provided by the authors and report the results according to standard evaluation protocols.

\noindent \textbf{Qualitative analysis.} We present visual comparisons of different methods on the SPIn-NeRF dataset, as shown in Figure \ref{fig:results_example1}. The first scene showcases the complex geometric structures of a staircase, which challenge the methods' ability to reconstruct its intricate details accurately. The results indicate that both SPIn-NeRF-based methods produce continuous but blurry areas that do not blend seamlessly with the surrounding scene. This limitation primarily stems from their heavy reliance on LaMa's inpainting results, which lack sufficient quality for complex geometric structures. Despite incorporating SDS priors, SPIn-NeRF-SDS cannot fully overcome the limitations inherited from LaMa's initial processing. While MVIP-NeRF generates smoother transitions, it struggles to maintain sharp geometric edges. In contrast, our method successfully reconstructs highly realistic geometric structures while preserving fine details.
The second scene presents a different challenge with an orange net, requiring accurate reproduction of its periodic texture. Among all methods, only MVIP-NeRF and our approach successfully reconstruct the net's repeated pattern. However, our method demonstrates superior performance by producing clearer and more realistic geometric details compared to MVIP-NeRF's results.

\noindent \textbf{Quantitative evaluation.} As shown in Table~\ref{tab:baseline}, our method achieves superior performance across most evaluation metrics. Specifically, our approach demonstrates excellent structural similarity and geometric faithfulness, as evidenced by high SSIM scores and favorable D-PSNR and D-FID metrics. Additionally, our method achieves outstanding visual quality, as indicated by strong performance in FID, NIMA, and BRISQUE metrics. While our method shows relatively lower performance in terms of PSNR, we argue that this metric alone is not sufficiently reliable for evaluating inpainting quality. This limitation of PSNR is particularly relevant given that inpainting is an inherently ill-posed problem~\cite{elharrouss2020image}, where multiple plausible solutions exist for unobserved regions. In such cases, the posterior mean solution usually produces the highest PSNR scores, but it often results in undesirably blurry outputs that compromise visual quality.

% \subsection{Analysis}
% \noindent \textbf{Ablation study.}
% To validate our approach, we perform ablation studies on two key components: the enhanced geometric priors in diffusion models and the improved score distillation method. As shown in Table~\ref{tab:ablation}, we achieve the best or the second best results in most metrics, and the suboptimal PSNR has been discussed before. Further, we also demonstrate the best visual quality in Fig.~\ref{fig:ablation}.
% Specifically, incorporating LoRA improves the model's ability to reconstruct the geometry better, as illustrated in Fig.~\ref{fig:ablation}. The model with LoRA achieved the highest BRISQUE and D-FID scores, indicating improved geometric performance after fine-tuning.
% Furthermore, the blurriness around the edges is reduced after adding BSD distillation. The model’s FID and NIMA scores also improve, suggesting that BSD helps reduce noise in the 3D model and enhances its consistency. Note that since the \textit{SPIN-NeRF} dataset has a separate test set, we use only the \textit{SPIN-NeRF} dataset for the ablation study metrics calculations.

% Furthermore, we explore alternative methods to fine-tune the diffusion model. (1) utilizing only the normal maps with the prompt ``normal map'' and (2) employing both normal maps and RGB images with prompts such as ``RGB image'' or ``normal map
% yielded suboptimal performance (as illustrated in Fig.~\ref{fig:normal_comp}). This observation underscores the importance of incorporating RGB images and extracting captions to fine-tune the diffusion model effectively.
\noindent \textbf{Ablation Studies.} We start by performing ablation studies to validate our key contributions: the enhanced geometric priors in diffusion models and the improved score distillation method. As demonstrated in Table~\ref{tab:ablation}, our complete method outperforms others across most metrics. In particular, by integrating LoRA for geometric prior learning, our approach significantly enhances the model's capability for geometric reconstruction, reducing D-FID by 25. Meanwhile, BSD consistently enhances visual quality by reducing optimization uncertainty, as evidenced by FID, NIMA, BRISQUE, and D-FID metrics improvements. Additionally, as discussed before, we ignore PSNR because it is not a reliable metric. By combining these two approaches, we can take advantage of their complementary benefits, resulting in the best overall performance. These quantitative improvements are also reflected in the visual results shown in Fig.~\ref{fig:ablation}, where LoRA notably improves geometric reconstruction accuracy, while BSD effectively reduces edge blurriness in the final results.
We further validate our strategy for fine-tuning the diffusion model. Our approach is based on two key insights: (1) it is essential to preserve the original appearance prior while also acquiring new geometric priors, and (2) utilizing Stable Diffusion's text comprehension capabilities is advantageous. To verify these insights, we conduct three comparative experiments: using only normal maps with the modality identifier ``normal map" as prompt; using both normal maps and RGB images with simple modality identifiers (``RGB image" or ``normal map") as prompts; and our approach of using both normal maps and RGB images with descriptive captions prepended with modality identifiers as prompts.

As shown in Fig.~\ref{fig:normal_comp}, alternative approaches yield suboptimal results, highlighting the importance of incorporating both RGB images and descriptive captions in the fine-tuning process.

\section{Conclusion}
In this work, we introduce GB-NeRF, a novel framework that enhances NeRF inpainting by more effectively utilizing 2D diffusion priors. First, we fine-tune the diffusion model to improve its ability to generate structurally accurate normal maps while still producing effective RGB images. Additionally, our proposed Balanced Score Distillation (BSD) technique outperforms existing methods, such as SDS and CSD, by delivering higher-quality inpainting with fewer artifacts. Our experiments demonstrate that GB-NeRF excels in both appearance fidelity and geometric consistency. However, our work has several limitations: (i) using fine-tuned diffusion priors for geometric enhancement increases training time; (ii) our method introduces new hyperparameters that require efforts to adjust; and (iii) similar to previous research \cite{mirzaei2023spin}, our method cannot eliminate shadows.

{
    \small
    \bibliographystyle{ieeenat_fullname}
    \bibliography{main}
}


\end{document}